# AN IMPROVED MULTIMODAL PSO METHOD BASED ON ELECTROSTATIC INTERACTION USING N-NEAREST-NEIGHBOR LOCAL SEARCH

Taymaz Rahkar-Farshi[1], Sara Behjat-Jamal[1], Mohammad-Reza Feizi-Derakhshi[2]

[1]Department of Computer Engineering, Gazi University, Ankara, Turkey
[2]Department of Computer Engineering, University of Tabriz,Tabriz, Iran

## ABSTRACT

*In this paper, an improved multimodal optimization (MMO) algorithm,calledLSEPSO,has been proposed. LSEPSO combinedElectrostatic Particle Swarm Optimization (EPSO) algorithm and a local search method and then madesome modification onthem. It has been shown to improve global and local optima finding ability of the algorithm. This algorithm useda modified local search to improve particle's personal best, which usedn-nearest-neighbour instead of nearest-neighbour. Then, by creating n new points among each particle and n nearest particles, it triedto find a point which could be the alternative of particle's personal best. This methodprevented particle's attenuation and following a specific particle by its neighbours. The performed tests on a number of benchmark functions clearly demonstratedthat the improved algorithm is able to solve MMO problems and outperform other tested algorithms in this article.*

## KEYWORDS

*Optimization; Multimodal Optimization; Particle Swarm Optimization; Evolutionary Computing.*

## 1. INTRODUCTION

One of the new issues in the field of evolutionary computing is MMO algorithms. Although various algorithms have been proposed in this field [1-4], most of these algorithms are based on PSO [5 – 11 ] and genetic [12 – 16 ] algorithms. The original forms of PSO and EA are designed for unimodal problems [17] and canonly locate a single global solution. In fact, all of the existing algorithms have been designed by modifying unimodal optimization algorithms.

In optimization problems, when there is a need to find more than one optimum, MMO algorithms are applied; practically, these algorithms have been only designed to find more than one optimum including local and global optima. In comparison with classic optimization algorithms, they have been designed only to fine one global optimum. Human needs lead to new problems, especially in engineering and management fields, and in order to solve these problems, new and various solutions are presented. Although unimodal optimization algorithms perform well in locating a single optimum, they cannot locate multiple optima. Problems such as clustering in machine learning and inversion of teleseismic waves [8 , 17 ] are some of those problems which can be solved by MMO algorithms; one of the most commonly used methods, based on which many studies have been conducted is niching method. In the presented algorithm, attempts were made to improve EPSO [9] by combining local search presented in [8] with its modification in order to improve the position of particle's personal best.

## 2. MULTIMODAL OPTIMIZATION

Most of real-world problems are in related to MMO. In fact, in such problems, there are more than one optimum and the aim is to find a set of optimal solutions. Since the structure and basis of classical optimization algorithms are only based on finding the best solution among the set of feasible solutions, it is obvious that such algorithms do not have the required capability in terms of solve these problems and the only feasible solution for these algorithms would be the global optimum; thus, local optimums not only cannot be the solutions, but also are obstacles to reaching the global optimum. In fact, these algorithms tend to converge quickly on an optimum solution [ 18 , 19 ]. In this respect, MMO algorithms have been applied to solve such problems. Although no algorithm has been designed to solve only these problems from the base thus far, recently, various algorithms have tried to solve these problems by modifying the existing classical optimization algorithms, a number of which will be introduced in the following section.

## 3. PSO ALGORITHM

The PSO algorithm inspired by the social behaviour of birds or fish, PSO algorithm is a swarm intelligence-based optimization algorithm which was presented by Kenedy and Eberhart in the mid-1990s [20]. This algorithm has been extensively welcomed owing to its more simple mechanism and very high efficiency; thus ,its implementation is much easier and simpler than that of other optimization algorithms. Basically, this algorithm has been designed to solve single-solution optimization problems; but, the mechanism of particles' motion in this algorithm has been designed in a way that, by some modification, it has been converted into a suitable algorithm in terms of solving MMO problems. This algorithm starts by distributing random particles in the problem space. For each particle, position, speed, and cost function values are considered and each particle has a memory in which the value and position of the best obtained solution, called personal best, are maintained by the particle itself. Also, there is a variable which maintains the value of cost function and position of the best solution obtained from all particles and is called global best. The new position of each particle is updated by its speed using Equation (1):

$$v_t = w.v_{t-1} + C_1.(p_i^{best} - x_i) + R_2.C_2.(g^{best} - x_i)$$
$$x_i = x_i + v_i$$
(1)

where $R_1$ and $R_2$ are two random variables within[0, 1], constants $C_1$ and $C_2$ are learning agents representing the attraction of each particle to its position or neighbours,parameter $C_1$ is a personal learning agent representing the attraction of each particle to its position, parameter $C_2$ is a global learning agent which represents the attraction of each particle to its neighbours, parameter $v$ is the speed which shows the direction and distance of a particle that must be traversed, and parameter $w$ is the inertia coefficient which controls speed. This algorithm normally is not able to find more than one optimum, since it is aimed to find a single solution[21].

## 4. RELATED WORKS

### 4.1. EPSO

EPSO algorithm has been proposed by J.Barrera and Carlos A.C. in 2009 [9]. In this algorithm, multimodal optimization problems are solved by modifying the mechanism of selecting global optimum in PSO algorithm. By applying Coulomb's law shown in Equation (2), the particles which should be separately selected as the global optimum for each particle are calculated. In fact, the particles may move toward different particles as global optimum; i.e. for each particle, the global optimum can be different, which leads to the fact that particles aggregate around local in addition to global optima. It is obvious that more particles aggregate around a point with a better cost function value. Feature of Coulomb's law is that particles tend to move toward a point which has suitable distance from that particle and a suitable cost function.

$$F_{(j,i)} = \frac{1}{4\pi\varepsilon_0} \cdot \frac{Q_1 \cdot Q_2}{r^2} \tag{2}$$

where F is the amount of electrostatic force, $Q_1$ and $Q_2$ are point charges, and $\frac{1}{4\pi\varepsilon_0}$ is called Coulomb's constant. Inspired by Coulomb's law, Equation (3) is obtained to calculate the force between two particles in PSO.

$$F_{(j,i)} = \alpha \cdot \frac{f(p_j) \cdot f(p_i)}{||p_j - p_i||^2} \tag{3}$$

$f(p_j)$ is the personal best value of the computing particle and $f(p_i)$ is the personal best value of a particle which has the possibility of moving toward it. The distance between particles $i$ and $j$ is computed in denominator. Value of $\alpha$ considered as Coulomb's constant and calculated by Equation (4) was presented in [7].

$$\alpha = \frac{||s||}{f(p_g) - f(p_w)} \tag{4}$$

where $||s||$ is the scaling factor which is assumed a problem space, $f(p_g)$ is the global optimum, and $f(p_w)$ is value of the worst cost function in the current population. As a result, particle $p_j$ would move toward a particle from $\vec{i}$ with the highest value F. At each iteration, the position and velocity of particle j are updated observing the following two rules in Equation (5).

$$v_t = w \cdot v_{t-1} + R_1 \cdot C_1 \cdot (p_i^{best} - x_i) + R_2 \cdot C_2 \cdot (p^F - x_i)$$
$$x_i = x_i + v_i \tag{5}$$

## 4.2 FERPSO

FERPSO [7] which was proposed by Xiaodong Li (2007) is a commonly used algorithm for solving MMO problems. Behavior of this algorithm can be described based on the nature view point: if there is more food in a place, more birds will aggregate there. In fact, if birds find suitable food around them, they would not go toward more resources at far points. In this algorithm, by applying Equation (2),the particles which are supposed to be selected as a global optimum for each individual particle are calculated. In fact, the general structure of FERPSO and EPSO is very similar and both have the same level of complexity.

$$FER_{(j,i)} = \alpha \cdot \frac{f(p_j) - f(\vec{x}_i)}{|p_j - x_i|} \tag{1}$$

$f(\vec{x}_i)$ is value of particle's computing cost function in the current population and $f(p_i)$ is the particle's personal best value which has the possibility of moving toward it. The dominator also computes the distance between the computing particle in the current population and other particles' personal best where there is the possibility of motion toward them. $\alpha$ is also computed as in Equation (4). As a result, particle $x_j$ would move toward particle $\vec{i}$ which has the highest value of FER. At each iteration, the position and velocity of $x_j$ are updated based on the following two rules in Equation (7).

$$v_t = w . v_{t-1} + R_1 . C_1 . (p_i^{best} - x_i) + R_2 . C_2 . (p^{FDR} - x_i) \qquad (7)$$
$$x_i = x_i + v_i$$

## 4.3 LSPSO

B.Y.Qu et al. [8] tried to solve such problems by combining a new local search technique with some multimodal PSO algorithms using niching method in 2012. In this method, by applying a local search, particle's personal best improves significantly. To achieve improvement, it generates a random point between the particle and the nearest neighbour; i.e. if the new point is better than the existing personal best, it will be replaced; otherwise, the previous value is not changed. This process is implemented by Equations(8, 9):

$$f(best\_nearest_i) \geq f(pbest_i) \rightarrow t \sum_{d=1}^{dim} p_{di}^{best} + C_1 . rand . (p_{di}^{best\_nearest} - p_{di}^{best}) \qquad (8)$$

$$f(best\_nearest_i) < f(pbest_i) \rightarrow t \sum_{d=1}^{dim} p_{di}^{best} + C_1 . rand . (p_{di}^{best} - p_{di}^{best\_nearest}) \qquad (9)$$

So, a new point would be generated between particle $i$ and its nearest neighbour; if $f(t)$ is better than $p_i^{best}$, it will replace $p_i^{best}$. Therefore, $p_i^{best}$ is updated and then combined with one of the algorithms using niching technique introduced in [5, 7, 21]. Although niching technique is commonly used in genetic algorithm to solve MMO problems, recently, various articles have been published for solving MMO problems by PSO using this technique [6]. A PSO algorithm in which niching method is used forsolving MMO problems was presented for the first time by Parsopoulos and Vrahatis [22, 23] to search for multiple global optima.

# 5.THE PROPOSED LSEPSO METHOD

In this paper, significant results were obtained for finding local and global optima by making some modification in the local search presented in [8] and combination with EPSO method presented in [9]. First, the personal best of each particle is improved by a new technique presented in Section 5.3 with the only difference that, instead of generating a new point between each particle's personal best and the nearest neighbour, n particles are generated among n near neighbours and the one with the best value is selected as an alternative candidate. Generation of the new point is done by Equation (8, 9). In fact, these equations would run n times in order to generate n points among the particle and n near neighbour. The important point is that, by increasing the value of n, particles are concentrated more on the optimal points with a higher value and the weaker optimal points have low chance in making particles concentrate on themselves. Even, the value of n can randomly change; so, by this method, particle's attenuation and also following a particular particle would be prevented. After updating personal best by combining EPSO algorithmthe final stage of the algorithm is done; the difference is that the value of α computed in Equation (4) is removed and α is set to 1. By iterating this process as approaching the end of the iteration, the particles are observed to aggregate around optimal points. By continuing the process, density of the particles would increase. Although performing these operations increases the complexity of algorithm, the results are improved over the normal case of EPSO algorithm and reach the optimum solution in the less number of iterations. The general pseudo-code of the algorithm is as follows:

---

**Algorithm 1** LSPSO

initialSwarm();
evaluate(Swarm); /* $\vec{x}$
**while** iteration
  local_search(k); k=(1:size(swarm))    /* **Algorithm 2**
  $F$ =EPSO (k);     k=(1:size(swarm))  /* **Algorithm 3**
  computeVelocities
  {
$v_{\text{iteration}} = w . v_{\text{iteration}-1} + R_1 . C_1 . (p_i^{best} - x_i) +$
$R_2 . C_2 . (p_{\square}^F - x_i);$
$x_{\text{iteration}} = x_{\text{iteration}} + v_{\text{iteration}};$
  }
  Updateposition();
  evaluateSwarm();
  update_best_position(); /* $p_{\square}^{best}$
**end while**

---

**Algorithm 2** Local_Search : Improve personal best for particle(i)

Input: index of particle = i
Dist(k)=distance(pbest(i),pbest(j)); j=(1:size(swarm))
n_best_nearst = n_min(dist);
best_nearst = max(evaluation(n_best_nearst));
**if** evaluate$(best\_nearest_i) \geq$ evaluate$(pbest_i)$
$t = \sum_{d=1}^{dim} p_{di}^{best} + C_1 . rand . (p_{di}^{best\_nearest} - p_{di}^{best})$
**Else**
$t = \sum_{d=1}^{dim} p_{di}^{best} + C_1 . rand . (p_{di}^{best} - p_{di}^{best\_nearest})$
**End if**
**If** evaluate (t)>evaluate(pbest(i))
  pbest(i)=temp;
**end**

---

**Algorithm 3** EPSO: Compute index maximum(i)

Input: index of particle = i
indmaxf = 1;
fmax = Inf;
Dist(k)=distance(pbest(i),pbest(j)); j=(1:size(swarm))
**for** j=1:size_of_swarm
**if** dist(j)>0
  F=pbest(i)*pbest(j)/(dist(j)^2);
  **if** j == 1
    fmax = F;
  **end if**
  **if** F>fmax
    fmax=F;
    indmaxf=j;
**end if**
**end if**
**end for**
return indmaxf;

## 6. TEST FUNCTIONS

The performed tests were done on common benchmark functions in MMO. Characteristics of these functions are listed in Table I. Since almost all the existing MMO algorithms perform well in one-dimension functions, applying tests on them was avoided and more challenging functions were considered. The test functions mentioned in Table I are as follows:

f1= Six-hump camel back, f2=Ackley, f3= Rastrigin, f4=Shubert, f5= Fifth function of De Jong

Table 1.  Test functions

| | Formula | Peaks: Global/Local | Range |
|---|---|---|---|
| $f1$ | $f(x_1, x_2) = \left(4 - 2.1x_1^2 + \dfrac{x_1^4}{3}\right) + x_1^2 + x_1x_2 + (-4 + 4x_2^2)x_2^2$ | 2/6 | $-1.9 < x_1 < 1.9$ $-1.1 < x_2 < 1.1$ |
| $f2$ | $f(x_1, x_2) = -20.\exp\left(-2.\sqrt{\frac{1}{2}(x_1{}^2 + x_2{}^2)}\right)\cdot$ $e^{\frac{1}{2}(\cos(cx_1) + \cos(cx_2))} + 20 + e$ | 1/121 | $-5 < x_1 < 5$ $-5 < x_2 < 5$ |
| $f3$ | $f(x_1, x_2) = 10n + \sum_{i=1}^{n} 10.2 + [x_i^2 - 10\cos(2\pi x)] + [y^2 - 10\cos(2\pi y)]$ | 1/121 | $-5.12 < x_1 < 5.12$ $-5.12 < x_2 < 5.12$ |
| $f4$ | $f(x_1, x_2) = -\sum_{i=1}^{5} i\cos(i+1) x_1 + 1] . \sum_{i=1}^{5} i\cos(i+1) x_2 + 1].$ | 4/201 | $-5.12 < x_1 < 5.12$ $-5.12 < x_2 < 5.12$ |
| $f5$ | $f(x_1, x_2) = \{0.002 + \sum_{j=1}^{25} \left[j + (x_1 - a_{1j})^6 + (x_2 - a_{2j})^6\right]^{-1}\}^{-1}$ $(a_{ij})$ $= \begin{pmatrix} -32 & -16 & 0 & 16 & 32 & -32 & 0 & 16 & 32 \\ -32 & -32 & -32 & -32 & -32 & -16 & \cdots 32 & 32 & 32 \end{pmatrix}$ | 1/36 | $-4 < x_1 < 4$ $-4 < x_2 < 4$ |

## 7. RESULTS AND COMPARISONS

The results of the performed tests are presented in Tables (2, 3). The first column in Table 2 represents the test functions, the second column is the number of particles, the third column is the number of iteration, and other columns show the average number of optima found(ANOF) per 10 executions for each algorithm. Table 3 represents ratio of the number of obtained optimum points to the number of existing optimum points in test functions. However, it must be noted that the obtained results of FERPSO algorithm were obtained with low accuracy and mean deviation of particles around optimal points was higher than those of other mentioned algorithms. The reason why the results were not mentioned for this algorithm in function $f4$ is higher scattering of the particles around optimal points and the optimal solution is not acceptable with this accuracy. Figure 1 shows the comparison of three FER-PSO, EPSO, and LSEPSO algorithms in 5 test functions.

As demonstrated by the results in the presented algorithm, it performed better than algorithm's normal state (EPSO) in all tests so that, in the performed comparisons in [9], this algorithm performed better than ANPSO[24], SPSO[5], and KPSO [25] algorithms. Also, comparison of the presented algorithm and FERPSO algorithm clearly showed that the former acted better than the latter. Figure 2 represents the search landscape of f5, respectively. Figure3 represents the position of particles during execution of the presented algorithm with 400 particles and the iteration number of 20 on f5 function. Figure 3.A to 3.E demonstrates the position of particles at1, 5, 10, and 20 iterations, respectively. Finally, Figure 3.F shows the obtained result after 20 iterations. In fact, 25 optimum points per 8000 times of executing optimum function were successfully obtained.

| Table I. Average number of optima found(ANOF) | | | | | |
|---|---|---|---|---|---|
| **Function** | **Particle** | **Iteration** | **FERPSO** | **EPSO** | **LSEPSO** |
| $f_1$= Six-hump camel back(2D) | 30 | 60 | 2.10 | 2.90 | **3.80** |
| | 60 | 60 | 2.30 | 4.30 | **4.50** |
| $f_2$= Ackley(2D) | 200 | 70 | 11.40 | 26.20 | **290** |
| | 400 | 70 | 11.50 | 46.30 | **53.50** |
| | 1000 | 50 | 17.50 | 81.20 | **94.40** |
| $f_3$= Rastrigin(2D) | 200 | 70 | 10.09 | 27.18 | **30.40** |
| | 400 | 80 | 17.63 | 54.10 | **60.70** |
| | 1000 | 60 | 22.72 | 88.00 | **106.50** |
| $f_4$= Shubert(2) | 200 | 70 | - | 26.50 | **31.10** |
| | 400 | 80 | - | 55.80 | **62.40** |
| | 1000 | 55 | - | 91.40 | **136.30** |
| $f_5$=Fifth function of De Jong(2D) | 200 | 30 | 9.40 | 19.40 | **21.11** |
| | 400 | 30 | 17.00 | 24.10 | **24.40** |

| Table 3. Average number of optima found/number of optima in test function | | | | | |
|---|---|---|---|---|---|
| **Function** | **Particle** | **Iteration** | **FERPSO** | **EPSO** | **LSEPSO** |
| $f_1$= Six-hump camel back(2D) | 30 | 60 | 0.35 | 0.483333 | **0.633333** |
| | 60 | 60 | 0.383333 | 0.716667 | **0.75** |
| $f_2$= Ackley(2D) | 200 | 70 | 0.094215 | 0.216529 | **0.239669** |
| | 400 | 70 | 0.095041 | 0.382645 | **0.442149** |
| | 1000 | 50 | 0.144628 | 0.671074 | **0.780165** |
| $f_3$= Rastrigin(2D) | 200 | 70 | 0.083388 | 0.224628 | **0.25124** |
| | 400 | 80 | 0.145702 | 0.447107 | **0.501653** |
| | 1000 | 60 | 0.187769 | 0.727273 | **0.880165** |
| $f_4$= Shubert(2) | 200 | 70 | - | 0.131841 | **0.154726** |
| | 400 | 80 | - | 0.277612 | **0.310448** |
| | 1000 | 55 | - | 0.454726 | **0.678109** |
| $f_5$=Fifth function of De Jong(2D) | 200 | 30 | 0.261 | 0.538 | **0.586** |
| | 400 | 30 | 0.472 | 0.666 | **0.6777** |

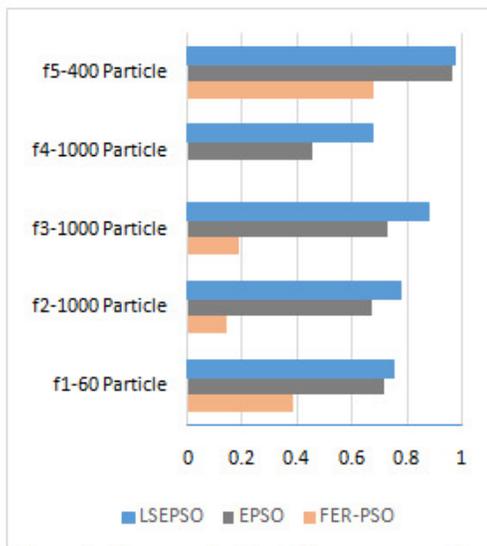

Figure 1. Diagram of ratio of the average number of optima found to the number of optima in test functions

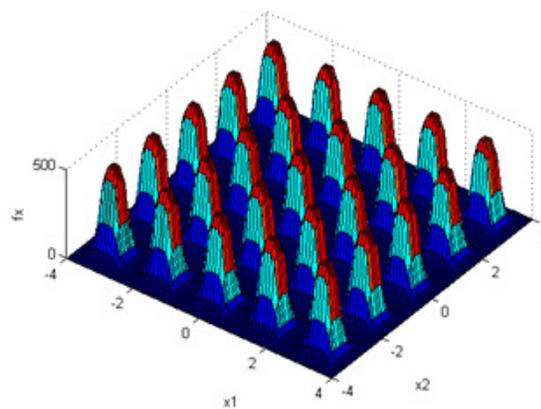

Figure 2. Search landscape of f5

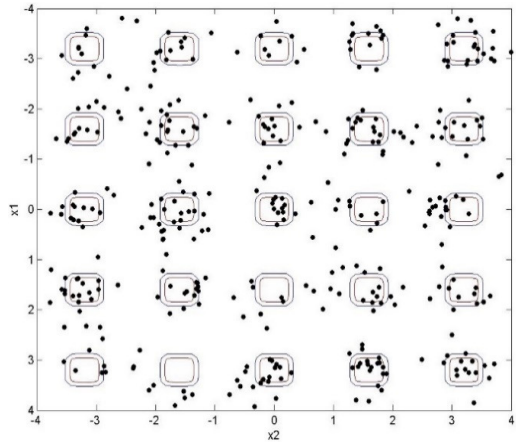
Figure3-A: Iteration 1

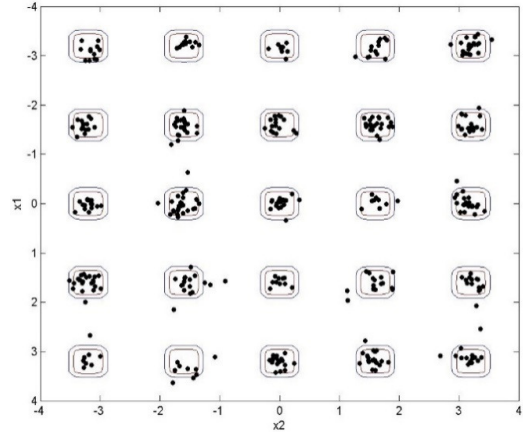
Figure 3-B: Iteration 5

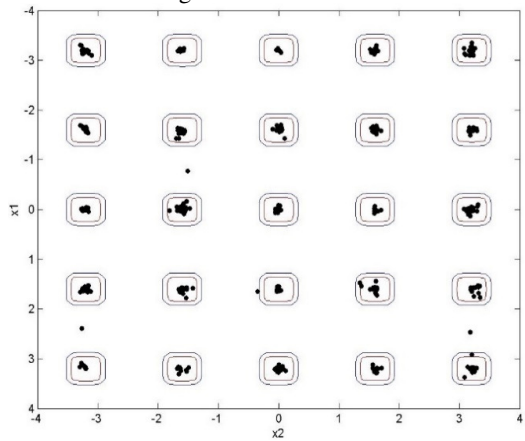
Figure 3-C: Iteration 10

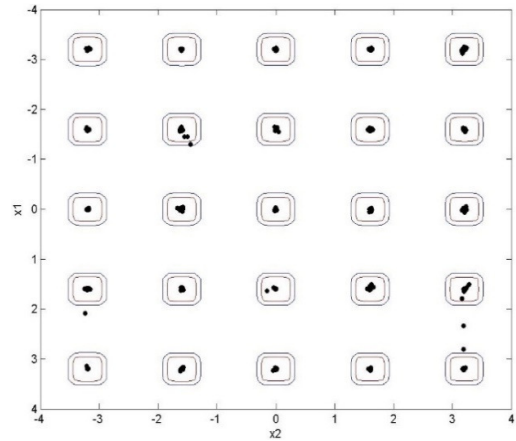
Figure 3-D: Iteration 15

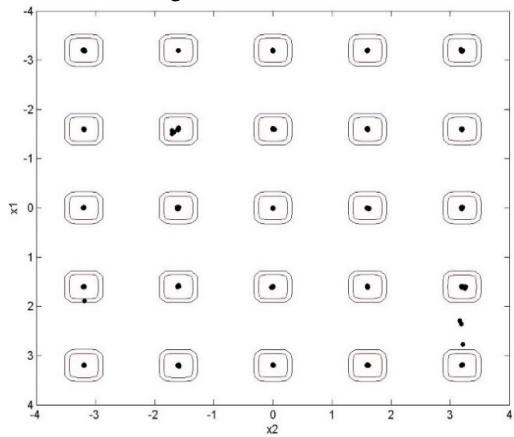
Figure 3-E: Iteration 20

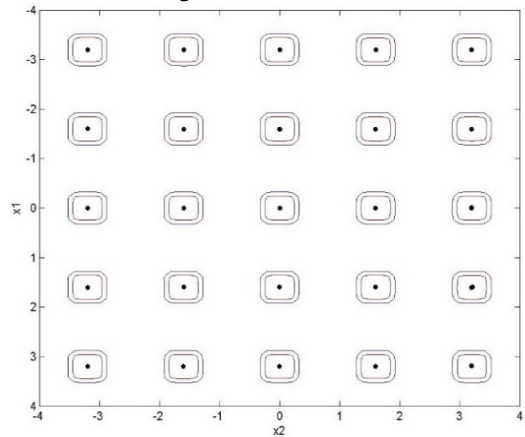
Figure 3-F: Result

Figure3.Position of particles at different iterations (beginning to end)

As demonstrated by the results in the presented algorithm, it performed better than algorithm's normal state (EPSO) in all tests so that, in the performed comparisons in [9], this algorithm performed better than ANPSO[24], SPSO [5] and KPSO [25] algorithms. Also, comparison of the presented algorithm and FERPSO algorithm clearly showed that the former acted better than the latter. Figure 2 represents the position of particles during execution of the presented algorithm with 400 particles and the iteration number of 20 on f5 function. Figures 3.A to 3.E demonstrate the position of particles at1, 5, 10, and 20 iterations, respectively. Finally, Figure 3.F shows the obtained result after 20 iterations. In fact, 25 optimum points per 8000 times of execution of optimum function were successfully obtained.

## 8. CONCLUSION

This article presented a new MMO algorithm (LSPSO) by combining a local search method and a MMO algorithm (EPSO) which has been proved to be more successful in solving MMO problems. The obtained results of the tests showed that this algorithm was more successful than EPSO algorithm's normal state and acted successfully in solving more challenging benchmark functions. One of the advantages of this algorithm was in the local search part, where by controlling the parameter n of the nearest neighbours, the particles around weak optimum points were removed and sent toward more powerful optimum points. Otherwise, when the weak optimum points were acceptable solutions as powerful optimum points, a low value was assigned to parameter n;so, the particles would not get away from the surrounding of weaker optimum points. As a result, it is obvious that this algorithm was a reliable algorithm in terms of solving MMO problems. It can be argued that all algorithms in this field have high time complexity which is one of the disadvantages of such algorithms and their execution speed is clearly less than normal optimization algorithms. One of the proper solutions for overcoming this problem could be the issue of parallelizing algorithms.

## REFERENCES

[1]  S. Yazdani, H. Nezamabadi-pour, and S. Kamyab, "A gravitational search algorithm for multimodal optimization," Swarm and Evolutionary Computation, vol. 14, pp. 1-14, 2014.

[2]  T. Rahkar-Farshi, O. Kesemen, and S. Behjat-Jamal, "Multi hyperbole detection on images using modified artificial bee colony (ABC) for multimodal function optimization," in Signal Processing and Communications Applications Conference (SIU), 2014 22nd, 2014, pp. 894-898.

[3]  A. Basak, S. Das, and K. C. Tan, "Multimodal Optimization Using a Biobjective Differential Evolution Algorithm Enhanced With Mean Distance-Based Selection," Evolutionary Computation, IEEE Transactions on, vol. 17, pp. 666-685, 2013.

[4]  B.-Y. Qu, P. N. Suganthan, and J.-J. Liang, "Differential evolution with neighborhood mutation for multimodal optimization," IEEE transactions on evolutionary computation, vol. 16, pp. 601-614, 2012.

[5]  X. Li, "Adaptively choosing neighbourhood bests using species in a particle swarm optimizer for multimodal function optimization," in Genetic and Evolutionary Computation–GECCO 2004, 2004, pp. 105-116.

[6]  M. Li, D. Lin, and J. Kou, "A hybrid niching PSO enhanced with recombination-replacement crowding strategy for multimodal function optimization," Applied Soft Computing, vol. 12, pp. 975-987, 2012.

[7]  X. Li, "A multimodal particle swarm optimizer based on fitness Euclidean-distance ratio," in Proceedings of the 9th annual conference on Genetic and evolutionary computation, 2007, pp. 78-85.

[8]  B.-Y. Qu, J. J. Liang, and P. N. Suganthan, "Niching particle swarm optimization with local search for multi-modal optimization," Information Sciences, vol. 197, pp. 131-143, 2012.

[9]  J. Barrera and C. A. C. Coello, "A particle swarm optimization method for multimodal optimization based on electrostatic interaction," in MICAI 2009: Advances in Artificial Intelligence, ed: Springer, 2009, pp. 622-632.

[10] B.-Y. Qu, P. N. Suganthan, and S. Das, "A distance-based locally informed particle swarm model for multimodal optimization," Evolutionary Computation, IEEE Transactions on, vol. 17, pp. 387-402, 2013.

[11] H. Wang, I. Moon, S. Yang, and D. Wang, "A memetic particle swarm optimization algorithm for multimodal optimization problems," Information Sciences, vol. 197, pp. 38-52, 2012.

[12] M. Li and J. Kou, "Crowding with nearest neighbors replacement for multiple species niching and building blocks preservation in binary multimodal functions optimization," Journal of Heuristics, vol. 14, pp. 243-270, 2008.

[13] T. Grüninger and D. Wallace, "Multimodal optimization using genetic algorithms," Master's thesis, Stuttgart University, 1996.

[14] R. K. Ursem, "Multinational GAs: Multimodal Optimization Techniques in Dynamic Environments," in GECCO, 2000, pp. 19-26.

[15] L. Wei and M. Zhao, "A niche hybrid genetic algorithm for global optimization of continuous multimodal functions," Applied Mathematics and Computation, vol. 160, pp. 649-661, 2005.

[16] E. Dilettoso and N. Salerno, "A self-adaptive niching genetic algorithm for multimodal optimization of electromagnetic devices," Magnetics, IEEE Transactions on, vol. 42, pp. 1203-1206, 2006.

[17] X. Li, "Niching without niching parameters: particle swarm optimization using a ring topology," Evolutionary Computation, IEEE Transactions on, vol. 14, pp. 150-169, 2010.

[18] P. Shelokar, P. Siarry, V. K. Jayaraman, and B. D. Kulkarni, "Particle swarm and ant colony algorithms hybridized for improved continuous optimization," Applied mathematics and computation, vol. 188, pp. 129-142, 2007.

[19] S. W. Mahfoud, "Niching methods for genetic algorithms," Urbana, vol. 51, 1995.

[20] K. James and E. Russell, "Particle swarm optimization," in Proceedings of 1995 IEEE International Conference on Neural Networks, 1995, pp. 1942-1948.

[21] J. Zhang, J.-R. Zhang, and K. Li, "A sequential niching technique for particle swarm optimization," in Advances in Intelligent Computing, ed: Springer, 2005, pp. 390-399.

[22] K. Parsopoulos and M. Vrahatis, "Modification of the particle swarm optimizer for locating all the global minima," in Artificial Neural Nets and Genetic Algorithms, 2001, pp. 324-327.

[23] K. E. Parsopoulos and M. N. Vrahatis, "On the computation of all global minimizers through particle swarm optimization," Evolutionary Computation, IEEE Transactions on, vol. 8, pp. 211-224, 2004.

[24] S. Bird and X. Li, "Adaptively choosing niching parameters in a PSO," in Proceedings of the 8th annual conference on Genetic and evolutionary computation, 2006, pp. 3-10.

[25] A. Passaro and A. Starita, "Particle swarm optimization for multimodal functions: a clustering approach," Journal of Artificial Evolution and Applications, vol. 2008, p. 8, 2008.

## AUTHORS

**Taymaz Rahkar-Farshi**
He was born in Tabriz, Iran, in 1985 and received his B.S. degree from University College of Nabi Akram in computer engineering. He received his M.S. degree in computer science from Karadeniz Technical University. He started his Ph.D. at Gazi University in 2013.

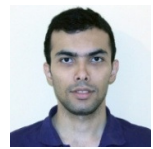

**Sara Behjat-Jamal**
She was born in Tabriz, Iran, in 1985. She received her B.S. from University College of Nabi Akram in computer engineering and her M.S. degree in computer engineering was from Gazi University.

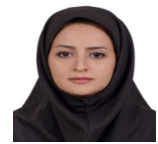

**Mohammad Reza Feizi-Derakhshi**
He received his B.S. in computer engineering, Department of Engineering, University of Isfahan, Isfahan, Iran. His M.S. and PhD degrees were from Department of Computer Engineering, Islamic Azad University- Science and Technology Branch, Tehran, Iran.

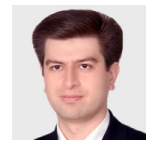